\documentclass[letterpaper, 10 pt, conference]{ieeeconf}  

\IEEEoverridecommandlockouts                              

\IEEEoverridecommandlockouts
\usepackage{cite}
\usepackage{amsmath,amssymb,amsfonts}
\usepackage{amsmath}
\usepackage{algorithm}
\usepackage{algpseudocode}

\usepackage{graphicx}
\usepackage{textcomp}
\usepackage{xcolor}
\usepackage{cuted}
\usepackage{subcaption}
\usepackage{booktabs}

\title{\LARGE \bf
Design of a Formation Control System to Assist Human Operators in Flying a Swarm of Robotic Blimps
}

\author{
Tianfu Wu$^{1}$, Jiaqi Fu$^{2}$, Wugang Meng$^{1}$, Sungjin Cho$^{3}$, Huanzhe Zhan$^{4}$ and Fumin Zhang$^{1,\dagger}$
\thanks{$^{\dagger}$\textbf{Corresponding author:} {\tt\small eefumin@ust.hk}}%
\thanks{*The work described in this paper was fully supported by grants AoE/E-601/24-N, 16203223, and C6029-23G from the Research Grants Council of the Hong Kong Special Administrative Region, China.}
\thanks{$^{1}$Tianfu Wu, Wugang Meng and Fumin Zhang are with the Department of Electronic and Computer Engineering, Hong Kong University of Science and Technology, Hong Kong.
        {\tt\small twubl@connect.ust.hk,wugang.meng@connect.ust.hk eefumin@ust.hk}}%
\thanks{$^{2}$Jiaqi Fu is with the School of Software Engineering, Beijing Jiaotong University, Beijing, China.
        {\tt\small liqifu964@gmail.com}}%
\thanks{$^{3}$Sungjin Cho is with the Department of Electronic Engineering, Sunchon National University, Suncheon, South Korea.
        {\tt\small sjcho@scnu.ac.kr}}%
\thanks{$^{4}$Huanzhe Zhan is with the Department of Computer Science, Emory University, Atlanta, GA 30322, USA.
        {\tt\small huanzhe.zhan@emory.edu}}%
}

\begin{document}

\maketitle
\thispagestyle{empty}
\pagestyle{empty}

\begin{abstract}
Formation control is essential for swarm robotics, enabling coordinated behavior in complex environments.  In this paper, we introduce a novel formation control system for an indoor blimp swarm using a specialized leader-follower approach enhanced with a dynamic leader-switching mechanism. This strategy allows any blimp to take on the leader role, distributing maneuvering demands across the swarm and enhancing overall formation stability. Only the leader blimp is manually controlled by a human operator, while follower blimps use onboard monocular cameras and a laser altimeter for relative position and altitude estimation.  A leader-switching scheme is proposed to assist the human operator to maintain stability of the swarm, especially when a sharp turn is performed. Experimental results confirm that the leader-switching mechanism effectively maintains stable formations and adapts to dynamic indoor environments while assisting human operator.
\end{abstract}

\section{INTRODUCTION}

Operating indoor blimp swarms for human-robot interaction opens up exciting opportunities across a wide range of applications, from surveillance to collaborative tasks in enclosed environments. Miniature Autonomous Blimps (MAB), compared to conventional drones, offer enhanced safety and extended operational times in human-centered spaces \cite{cho2022autopilot,van2000vision,hou2022human}. However, simultaneously controlling multiple blimps presents significant challenges, especially for a single operator.
Current systems rely heavily on external localization methods such as OptiTrack, which, although providing precise indoor positioning, are expensive and face challenges with scalability and deployment flexibility. These limitations restrict the autonomy and adaptability of blimp swarms in various applications.

Although current swarm control methods, such as behavior-based control, consensus algorithms, and potential field methods, do not require external localization systems, they still rely on multiple onboard sensors (e.g., cameras, LiDAR) for decentralized coordination \cite{Gaofei,zhou2023racer,zhou2021ego,xu2022omni}. However, when applied to blimps, these approaches face significant challenges due to the limited payload capacity and onboard computational power.
\begin{figure}[t]
    \centerline{\includegraphics[width=0.45\textwidth]{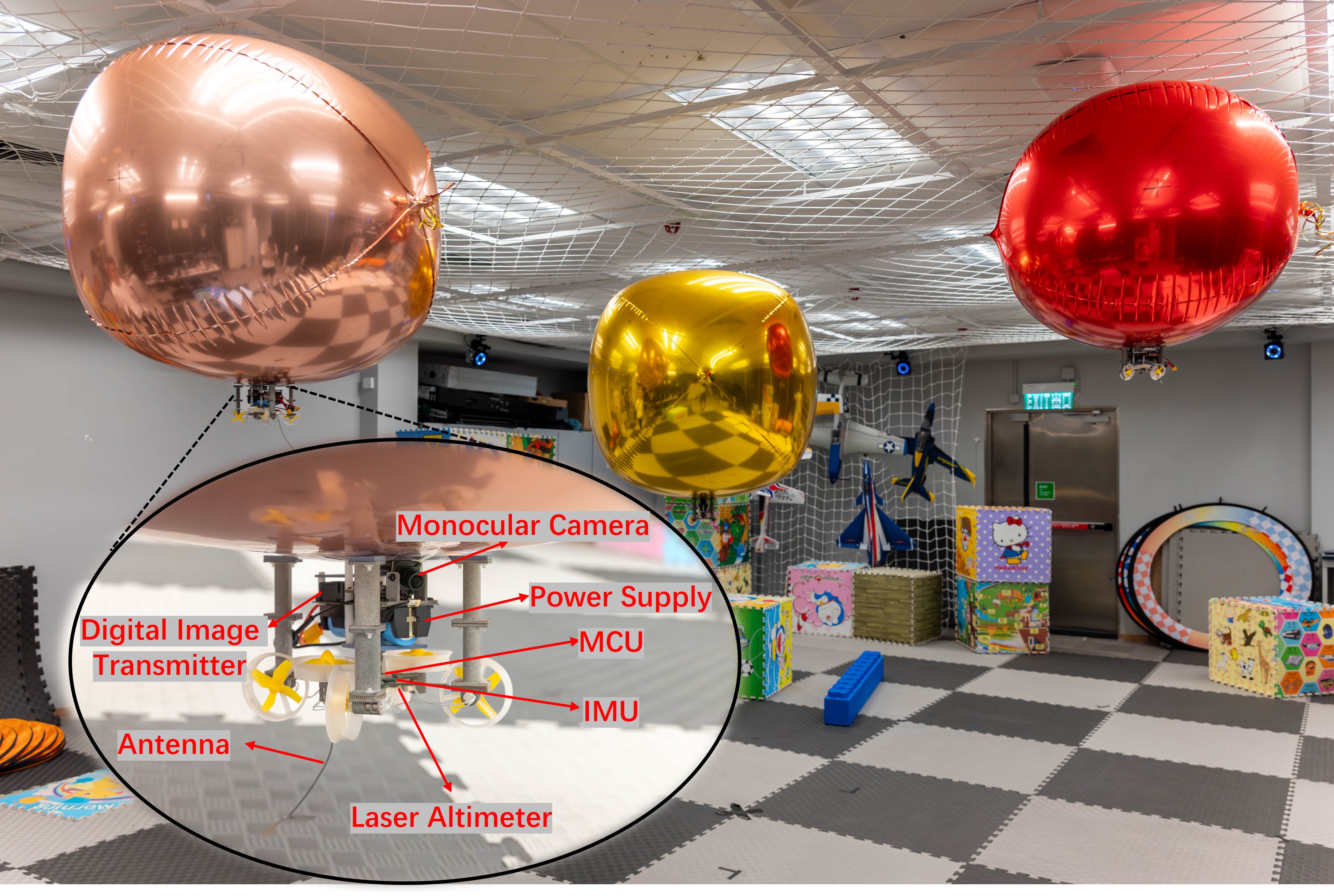}}
    \caption{Indoor blimp swarm and the blimp design. }
    \label{fig:blimp swarm}
\end{figure}
Due to payload constraints, blimps cannot be equipped with advanced localization tools like UWB modules or multiple cameras for accurate positioning, relying instead on a single monocular camera. This limitation can lead to issues during maneuvers, as one follower blimp may block another's field of view, disrupting the formation and increasing the operator's workload. Additionally, the traditional leader-follower strategy is complicated by the blimps' underactuated nature and low maneuverability, making precise control difficult, especially during turns. These challenges often result in transient motions that further disrupt formations in indoor environments, where maintaining consistency is critical \cite{Qiuyangthesis,yao2017monocular}.

Despite recent progress in swarm control techniques, achieving efficient management of blimp swarms with minimal operator input continues to pose significant challenges. This underscores the need for innovative approaches that not only enhance human-swarm interaction but also incorporate the operator more seamlessly into the control loop, allowing for smoother and more intuitive swarm management.

To address these challenges, we propose a novel control system for indoor blimp swarms, enhancing the traditional leader-follower approach with a dynamic leader-switching algorithm. This algorithm allows any blimp to dynamically assume the leader role, distributing maneuvering demands across the swarm and reducing disruptions during complex maneuvers. The system also introduces a leader-search capability for followers that lose sight of the leader, allowing them to autonomously reconnect without constant operator intervention. To manage the computational limitations of blimps, visual data processing is offloaded to an external computer. 
Our contributions are summarized as follows:
\begin{enumerate}
    \item \textbf{A Novel Blimp Control System:} We introduce a control system for indoor blimps that minimizes onboard sensors and offloads computational tasks to external computers, enabling rapid deployment without external localization devices.
    \item \textbf{Dynamic Leader-Switching Algorithm:} We develop a leader-switching algorithm that dynamically adjusts the leader role within the swarm, improving coordination and stability.
    \item \textbf{Enhanced Human-Robot Swarm Interaction:} We design an interface that reduces operator workload by providing real-time monitoring of blimp camera feeds and flight data, along with automated error detection to assist in managing the swarm.
\end{enumerate}

\section{RELATED WORK}

\subsection{Traditional Leader Follower Strategy}
Leader-follower strategies are a foundational approach in multi-robot coordination, widely used for formation control in both ground and aerial robots. These methods typically designate a leader to dictate the movement trajectory, while followers maintain relative positions, ensuring coordinated behavior. Basic geometric approaches using visual servo methods have been applied successfully in maintaining formations with simple sensors like fisheye cameras \cite{oh2023leader}. More advanced algorithms, such as dynamic priority strategies and receding horizon schemes, enhance adaptability by dynamically updating paths based on environmental changes \cite{liu2013leader}. However, these methods often assume reliable localization and high-frequency communication \cite{jia2018distributed}, which are challenging for blimp swarms due to constraints like limited payload, low maneuverability, and reduced computational power.

\begin{figure}[h]
    \centerline{\includegraphics[width=0.3\textwidth]{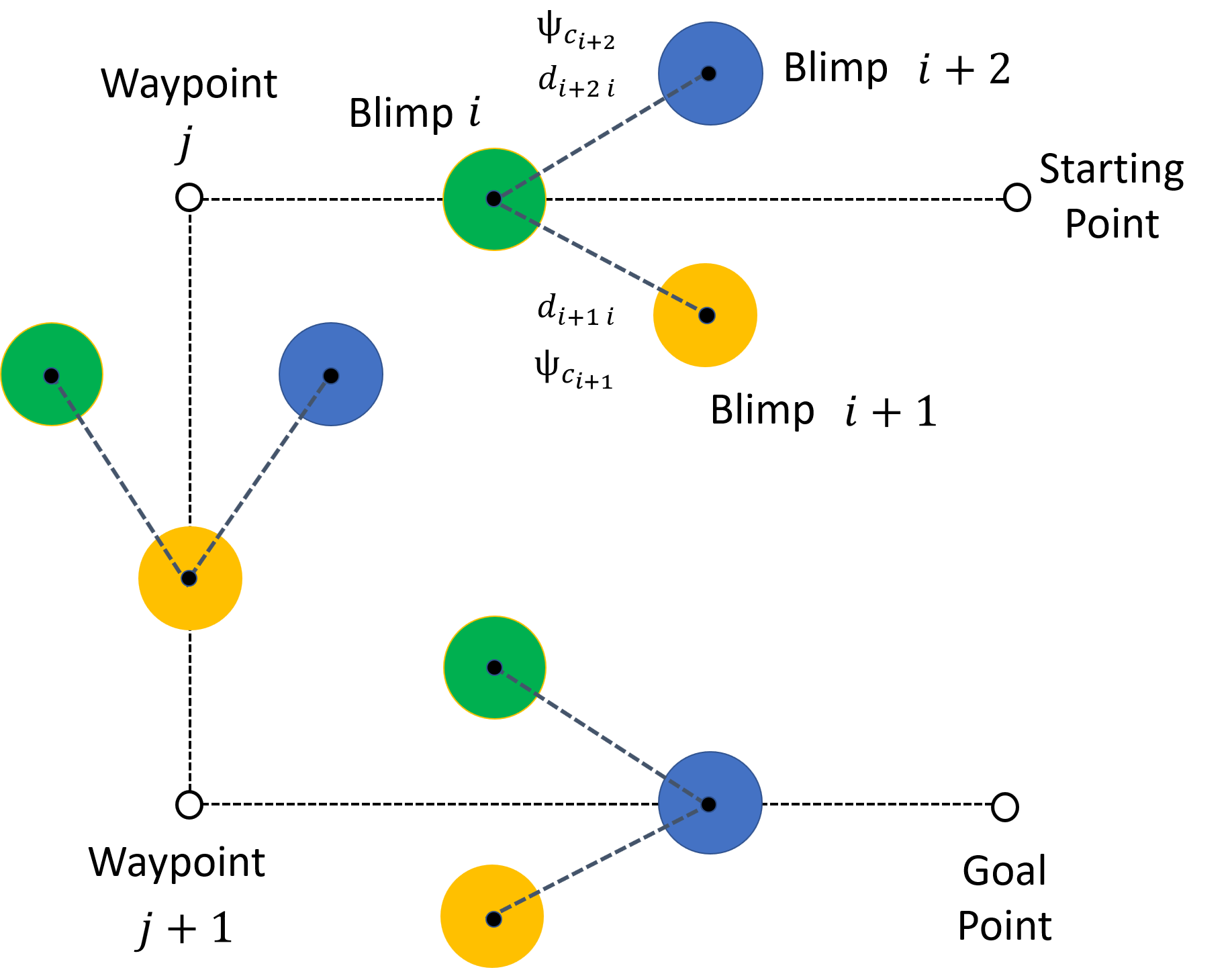}}
    \caption{Illustration of operator-controlled leader-switching during sharp turns.}
    \label{fig:pic3}
\end{figure}

\subsection{Dynamic Leader Selection}

Dynamic leader selection has been widely explored in multi-robot systems to enhance flexibility and adaptability, using criteria such as battery levels, sensor data, or environmental feedback. For instance, Li et al. adjusted leader roles based on proximity to a target \cite{li2015dynamic}, and Dotson et al. introduced an energy-efficient leader-switching policy inspired by bird flocking \cite{dotson2019energy}. Other approaches include Delkhosh's decentralized coordination model for emergencies \cite{delkhosh2020dynamic}, Swaminathan's failure-prevention strategy in cluttered environments \cite{swaminathan2015planning}, and Xiao's use of model predictive control to stabilize leader-follower dynamics via neural-dynamic optimization \cite{xiao2016formation}.
However, these methods depend on autonomous operation and high maneuverability, which are unsuitable for blimps due to their low agility, communication constraints, and limited computational power. Blimps struggle with coordination, especially under unpredictable conditions requiring real-time decisions.

Our approach involves a human operator for high-level decision-making, compensating for these limitations and reducing the risk of formation loss. It features dynamic leader-switching, detecting when a follower loses sight of the leader or an incorrect leader is chosen, thereby minimizing operator workload and enhancing system adaptability. This hybrid method improves stability and responsiveness in indoor environments.

\begin{figure*}[t]
    \centering
    \includegraphics[width=0.7\textwidth]{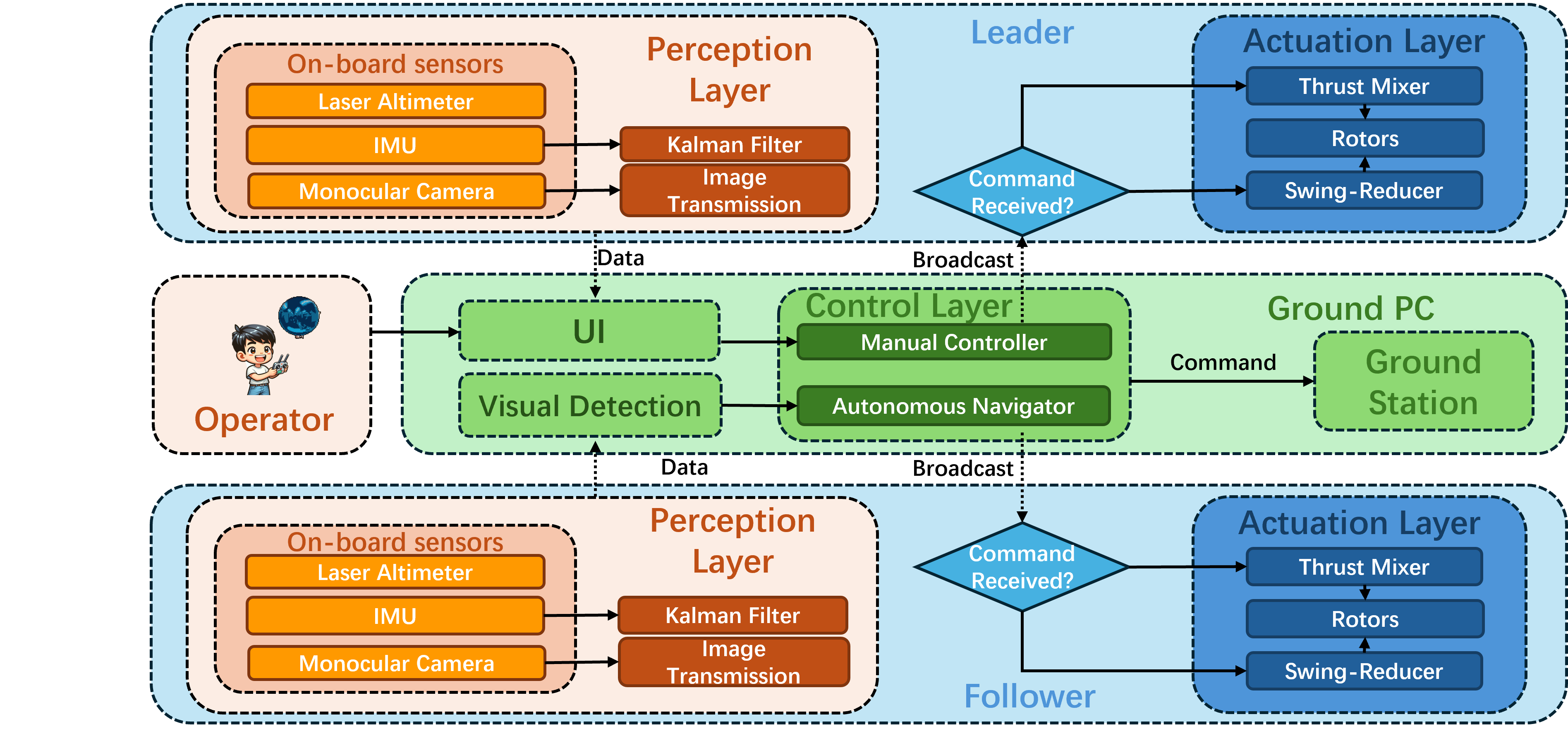} 
    \caption{Block diagram of indoor aerial swarm formation flight with three blimps.}
    \label{fig:system}
\end{figure*}

\subsection{Problem Formulation}
Maintaining formation control for a swarm of indoor blimps is challenging due to their limited payload, sensing capabilities, and underactuated dynamics, especially in cluttered environments. Sharp turns pose a significant issue as the tight coupling between translational and rotational movements can cause sudden deviations in trajectories\cite{Qiuyangthesis}, leading followers to lose sight of the leader and disrupt the formation. These challenges are exacerbated by the blimps' limited sensing capabilities, increasing the risk of formation breakdowns and necessitating frequent manual corrections, which burden the operator and reduce system efficiency.

To address these issues, we propose a control system with a dynamic leader-switching algorithm managed via a custom user interface (UI). The operator monitors real-time video feeds and dynamically assigns the leader role among the three blimps. By switching leaders during sharp turns, the system distributes maneuvering demands, reduces motion intensity, and helps maintain formation stability. Figure \ref{fig:pic3} illustrates the leader-switching process, showing how the system transitions between leaders to stabilize the formation.

The system ensures safe leader switching by verifying that the new and current leaders are within each other's visual range before executing the switch, facilitating smooth transitions. After the switch, the remaining blimps automatically follow the new leader, minimizing the need for manual adjustments. If an incorrect leader is selected, the system alerts the operator to reduce errors in leader selection. This approach alleviates the operator's workload, enhances formation stability, and improves the swarm's adaptability in dynamic indoor environments, making the system more effective and reliable.

\section{METHODOLOGY}
\subsection{System Overview}
Fig. \ref{fig:system} illustrates the proposed system, which consists of leader and follower, each with perception, control, and actuation layers. These agents collaborate through a Ground PC, which manages communication and coordination. The system allows dynamic switching between the leader and followers, enabling any agent to take the leader role as needed. Below is a summary of the components

\begin{itemize}
    \item \textbf{Perception Layer}: 
    Each agent is equipped with a laser altimeter, IMU, and monocular Camera. Sensor data is processed via a Kalman Filter, and the followers use visual detection to track the leader.

    \item \textbf{Control Layer}:
    The leader is manually controlled by the operator, while the followers autonomously track the leader using visual input. The system supports dynamic role switching, coordinated by the Ground PC.

    \item \textbf{Actuation Layer}:
    This layer includes a Thrust Mixer, Rotors, and Swing-Reducer for movement control. Followers adjust their actuators based on visual input to maintain formation.

    \item \textbf{Communication and Coordination}: 
    The Ground PC broadcasts control signals. The operator controls the leader, and the followers operate autonomously, with the ability to switch roles as needed.
\end{itemize}

This system ensures coordinated operation between the leader and followers, with flexible role switching for robustness in complex environments.

\subsection{Leader Switch Algorithm}
The leader switch algorithm enables the current leader blimp to transfer leadership to a new blimp. When the current leader blimp (the $i^{th}$ blimp) reaches the $j^{th}$ waypoint like in Fig. \ref{fig:pic3}, the human operator checks if the next waypoint (the $j+1^{th}$ waypoint) is perpendicular to the $j^{th}$ waypoint. If the next waypoint is located to the left of the current one, the $i+1^{th}$ blimp is chosen as the new leader. If it is on the right, the $i+2^{th}$ blimp becomes the new leader. 
\begin{figure}[t]
    \centerline{\includegraphics[width=0.45\textwidth]{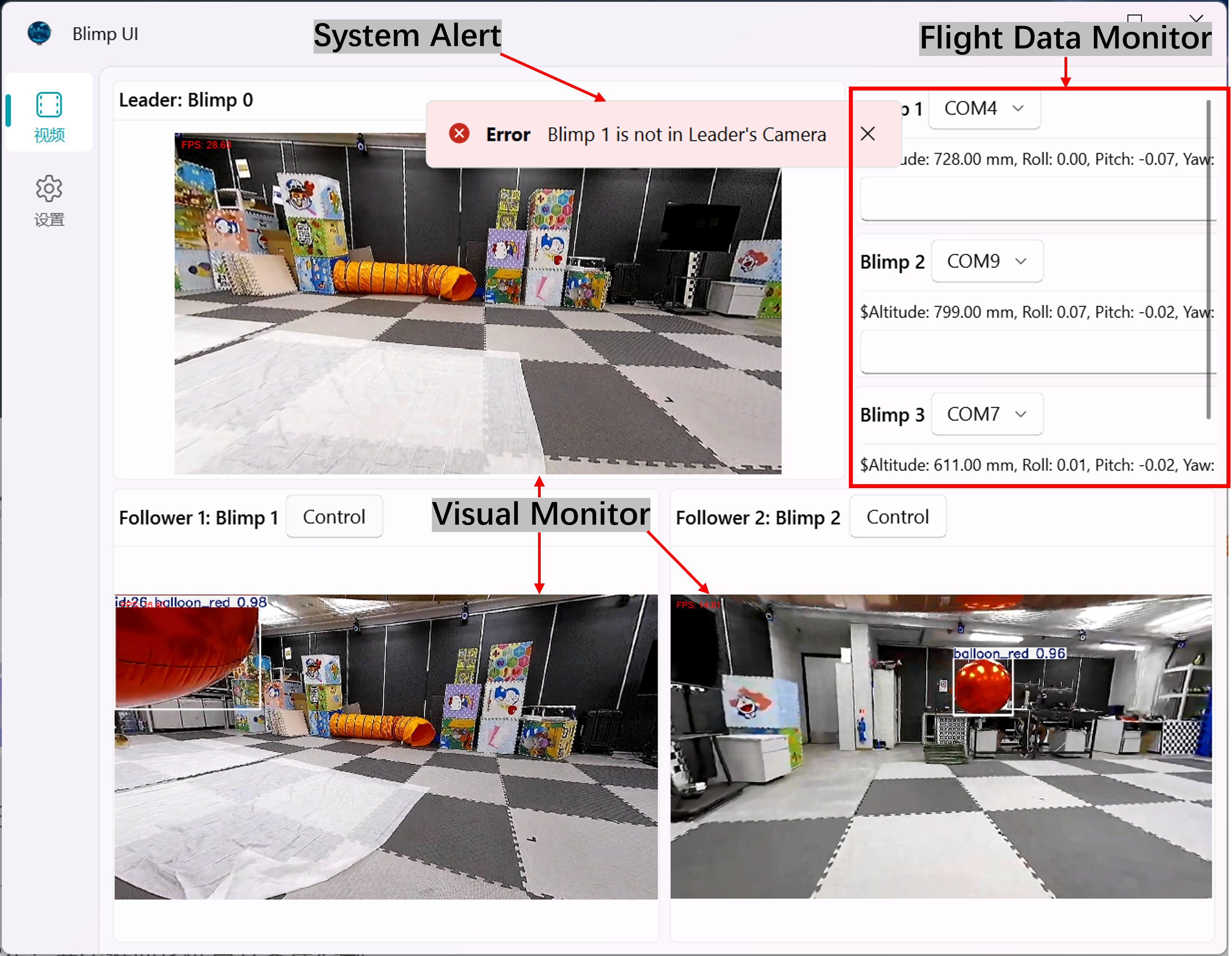}}
    \caption{Blimp swarm control UI with leader selection error alert.}
    \label{fig:UI}
\end{figure}

To ensure a smooth leader switch, both the current and new leader blimps must maintain mutual visibility. This visual connection allows them to accurately estimate each other's relative positions and orientations, facilitating an efficient leadership transition. If the current leader does not see the new leader within its field of view, an error will be triggered on the user interface (UI), as shown in Fig. \ref{fig:UI}. At this point, the operator must manually rotate the leader blimp using the UI to establish visual contact with the new leader. The UI provides real-time feedback on each blimp’s field of view, enabling the operator to make precise adjustments and ensure the leader switch occurs smoothly.
Meanwhile, the other follower blimp, which is not involved in the switch, will automatically begin scanning for the new leader. It does not need to maintain mutual visibility with the new leader, but only needs to detect the new leader within its own field of view. If the follower detects the new leader, it will immediately enter follow mode without further searching. However, if the new leader is not initially visible, the follower will continue its rotational search until it establishes visual contact with the new leader, and then it will transition seamlessly into follow mode.

dThis process ensures that the entire system remains coordinated and responsive during the leader switch. The operator intervenes only when the current leader cannot see the new leader, while the remaining followers autonomously adjust their behavior to maintain formation. This approach minimizes interruptions in the swarm's movement and enhances the system’s adaptability during dynamic operations.

\begin{figure}[t]
    \centerline{\includegraphics[width=0.4\textwidth]{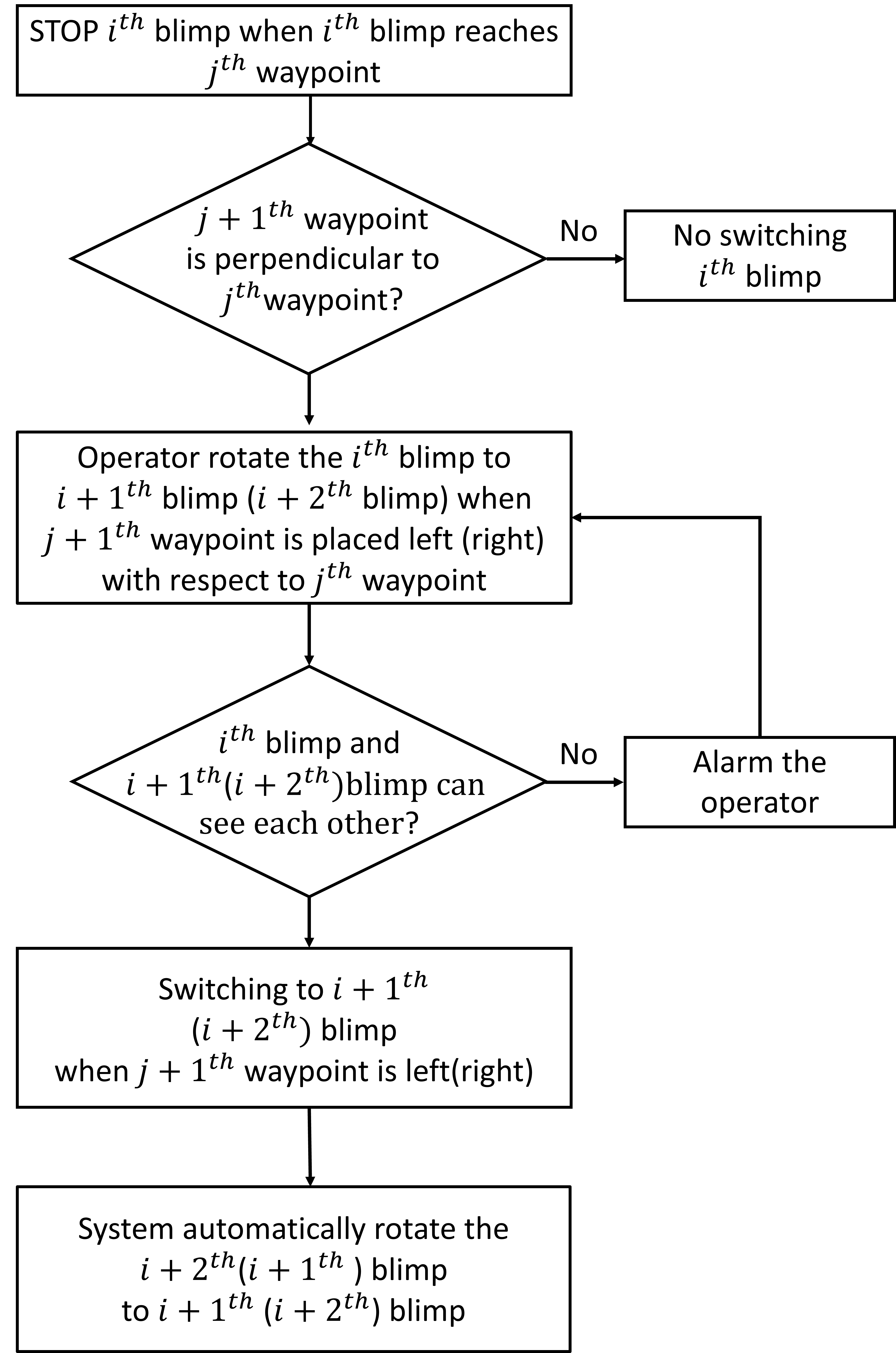}}
    \caption{Flow Charts of Leader Switch Algorithm}
    \label{fig:pic5}
\end{figure}

\subsection{Monocular Camera Relative Position Estimation}

Due to the blimp's limited payload capacity, we utilize a standard monocular camera for relative position estimation, as opposed to systems that use multiple cameras, depth sensors, or positioning systems like OptiTrack. In previous work \cite{yao2017monocular}, monocular cameras were used for position estimation, but the inability to directly measure depth required complex 3D inference algorithms, leading to potential errors in dynamic environments. To address this, we integrated a laser altimeter to measure altitude directly, reducing reliance on depth estimation and improving real-time accuracy in challenging scenarios.

\begin{figure}[h]
    \centerline{\includegraphics[width=0.35\textwidth]{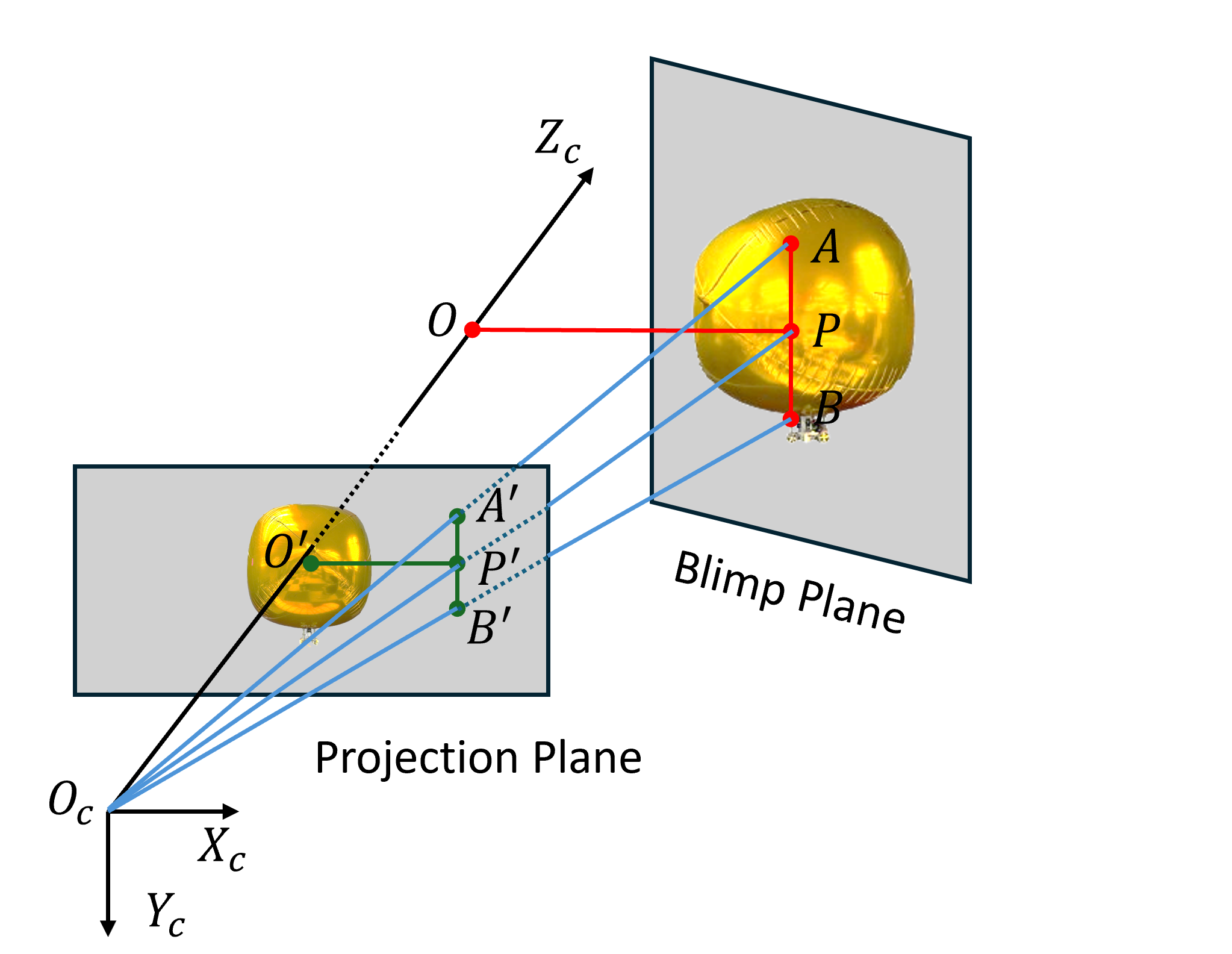}}
    \caption{Leader blimp relative position estimation from the follower’s perspective.}
    \label{fig:Distance}
\end{figure}

As shown in Fig. \ref{fig:Distance}, the follower blimp estimates the leader blimp’s position assuming that the camera's projection plane remains perpendicular to the ground due to minimal pitch and roll. The leader blimp’s centerline \(AB\) is parallel to the image plane, with its center \(P = \begin{bmatrix} x_P, y_P, z_P \end{bmatrix}^T\) projected to the image coordinates \(P' = \begin{bmatrix} i_P, j_P \end{bmatrix}^T\). Let \(L_0 = |AB|\) represent the actual length of the leader blimp, and \(l_f = |A'B'|\) be its observed length in the image.
At a known distance \(d_0\) from the camera, the blimp’s projected length in the image is measured as \(l^0_f\), and the focal length \(f\) is calculated as:

\begin{equation}
f = d_0 \frac{l^0_f}{L_0}
\end{equation}

Once the focal length \(f\) is determined, for any observed image length \(l_f\), the center coordinates of the leader blimp \(\hat{x}_P, \hat{y}_P, \hat{z}_P\) can be estimated as follows:

\begin{align}
\hat{z}_P &= d_0 \cdot \frac{l^0_f \cdot L_0}{L_0 \cdot l_f} = d_0 \cdot \frac{l^0_f}{l_f}, \notag \\
\hat{x}_P &= \hat{z}_P \cdot \frac{(i_P - i_0)}{f} = \frac{(i_P - i_0)}{l^0_f \cdot d_0 / L_0} \cdot d_0 \cdot \frac{l^0_f}{l_f} = L_0 \cdot \frac{(i_P - i_0)}{l_f}, \notag \\
\hat{y}_P &= \hat{z}_P \cdot \frac{(j_P - j_0)}{f} = \frac{(j_P - j_0)}{l^0_f \cdot d_0 / L_0} \cdot d_0 \cdot \frac{l^0_f}{l_f} = L_0 \cdot \frac{(j_P - j_0)}{l_f}. \label{eq:combined}
\end{align}

In these equations:
\(\hat{z}_P\) is the depth (or distance) of the leader from the \(\hat{x}_P\) and \(\hat{y}_P\) are the horizontal and vertical positions of the leader blimp relative to the follower’s camera.

Finally, the relative distance \(\hat{d}\) and yaw angle \(\hat{\psi}\) of the leader blimp are derived as:

\begin{equation}
\hat{d} = \sqrt{\hat{x}_P^2 + \hat{z}_P^2}, \quad \hat{\psi} = \arcsin\left( \frac{\hat{x}_P}{\hat{d}} \right)
\end{equation}

This estimation allows the follower blimp to track the leader blimp’s relative position and orientation in real time, ensuring precise formation control.

\subsection{Follower Controller Design}
The blimp's motion exhibits significant nonlinearities and strong coupling between translational and rotational movements. Without external positioning devices or a velocity sensor, direct velocity control for waypoint tracking is not feasible. However, in our previous study \cite{CCDC}, we developed a set of balanced controllers that leverage the coupling between horizontal velocity and pitch angle to achieve effective velocity tracking.

\begin{figure}[h]
    \centerline{\includegraphics[width=0.45\textwidth]{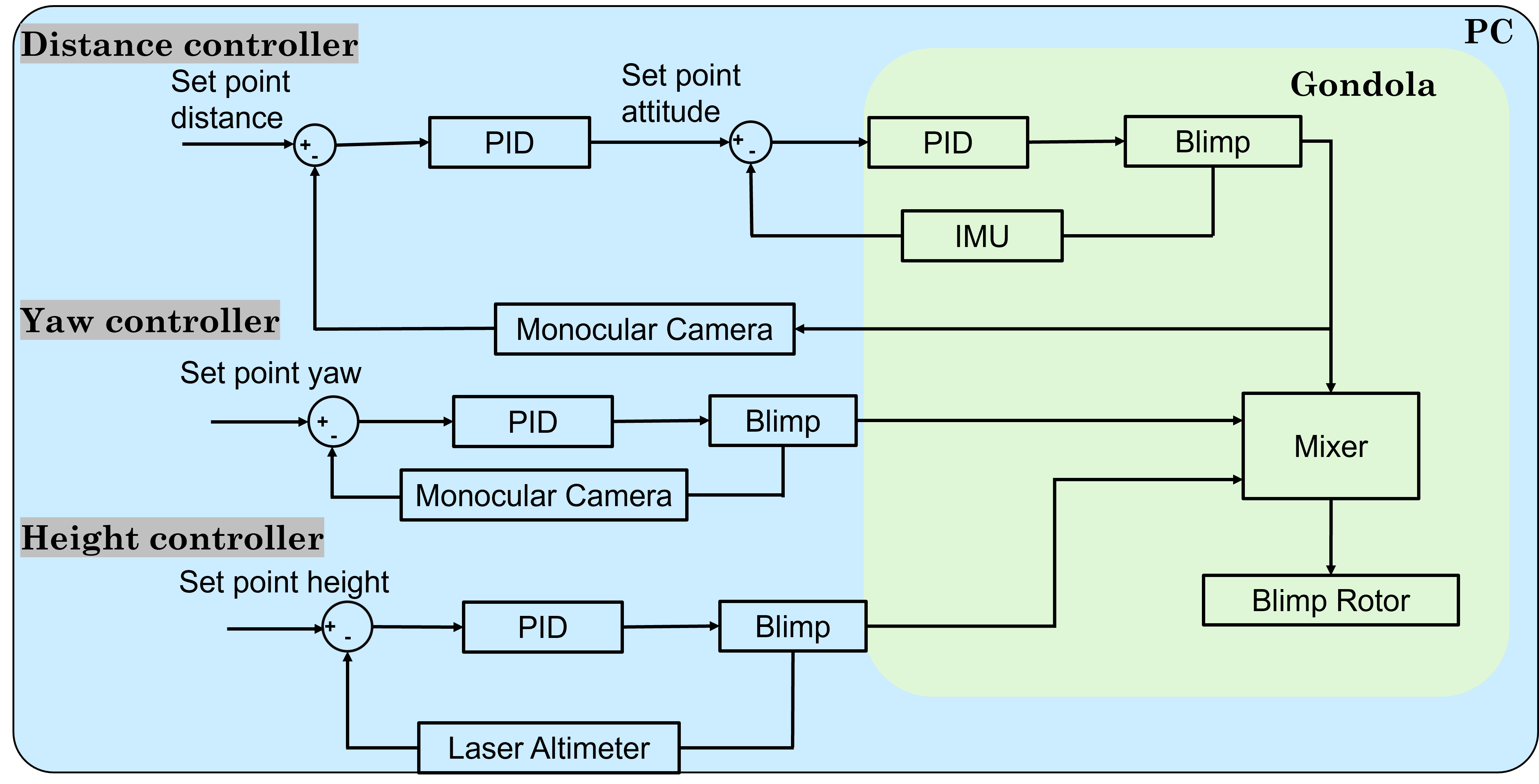}}
    \caption{Block diagram of the follower controller.}
    \label{fig:Controller}
\end{figure}

\subsubsection{Distance Controller}

The distance controller adjusts the follower blimp's pitch angle to regulate the horizontal distance from the leader. As derived in \cite{CCDC}, the horizontal acceleration \(\dot{v}_h\) is coupled with the pitch angle \(\theta\) according to the following relationship:

\begin{equation}
\dot{v}_h = \cos{\theta} \left(\dot{v}_0 + \frac{F_h}{m}\right)
\end{equation}

where \(\dot{v}_h\) is the horizontal acceleration, \(\theta\) is the pitch angle, \(F_h\) is the applied thrust, and \(\dot{v}_0\) is the acceleration due to air resistance, which is particularly small and can be neglected. To ensure reliable distance estimation from the monocular camera, the pitch angle \(\theta\) is kept within a limited range, preventing excessive tilting that could compromise the camera's accuracy in measuring the distance to the leader. Since \(\theta\) remains small, we can approximate \(\cos{\theta} \approx 1\)\cite{CCDC}.

\begin{equation}
\begin{split}
{v}_h = & K_{p_1} \left( d_{\text{setpoint}} - \hat{d} \right) \\
            & + K_{i_1} \int_0^t \left( d_{\text{setpoint}} - \hat{d}(\tau) \right) d\tau \\
            & + K_{d_1} \frac{d}{dt} \left( d_{\text{setpoint}} - \hat{d}(t) \right)
\end{split}
\end{equation}
\begin{equation}
\begin{split}
\theta = & K_{p_2} \left( {v}_h - {v}_{h,\text{current}} \right) \\
         & + K_{i_2} \int_0^t \left( {v}_h - {v}_{h,\text{current}}(\tau) \right) d\tau \\
         & + K_{d_2} \frac{d}{dt} \left( {v}_h - {v}_{h,\text{current}}(t) \right)
\end{split}
\end{equation}

The distance controller uses a PID control loop to adjust the follower blimp's pitch angle \(\theta\), with feedback provided by the monocular camera for distance estimation \(\hat{d}\) and the IMU for pitch angle measurements \(\theta\). The outer PID loop manages the distance error between the setpoint \(d_{\text{setpoint}} = 1.5\,\text{m}\) and the current estimated distance \(\hat{d}\), while the inner PID loop fine-tunes the pitch angle \(\theta\) to achieve the desired horizontal velocity \({v}_h\).

To ensure reliable distance estimation from the monocular camera, the pitch angle \(\theta\) is kept within a limited range, preventing excessive tilting that could compromise the camera's accuracy in measuring the distance to the leader.





\subsubsection{Height Controller}
In our blimp design, vertical thrust is generated by two vertical rotors. The altitude control system uses a single-loop PID controller to adjust the vertical thrust \(f_y\) based on the error between the desired altitude \(h_{\text{setpoint}}\) and the current altitude \(h(t)\), with feedback provided by a laser altimeter.
\begin{equation}
\begin{split}
f_y(t) = & K_p \left( h_{\text{setpoint}} - h(t) \right) \\
         & + K_i \int_0^t \left( h_{\text{setpoint}} - h(\tau) \right) d\tau \\
         & + K_d \frac{d}{dt} \left( h_{\text{setpoint}} - h(t) \right)
\end{split}
\end{equation}




\subsubsection{Yaw Controller}
The yaw controller regulates the blimp's rotational motion around its vertical axis to achieve the desired yaw angle \(\psi_{\text{setpoint}} = 0\). It uses a PID control loop that adjusts the yaw torque \(\tau\) based on the error between the desired yaw angle and the current yaw angle \(\hat{\psi}\), with feedback provided by the monocular camera. To maintain accurate yaw control and prevent excessive rotations that could affect the monocular camera's performance, the system carefully limits yaw adjustments, ensuring that the blimp stays properly oriented within the formation.

\begin{equation}
\begin{aligned}
\tau(t) = & \, K_p \left( \psi_{\text{setpoint}} - \hat{\psi}(t) \right) \\
            & + K_i \int_0^t \left( \psi_{\text{setpoint}} - \hat{\psi}(\tau) \right) d\tau \\
            & + K_d \frac{d}{dt} \left( \psi_{\text{setpoint}} - \hat{\psi}(t) \right)
\end{aligned}
\end{equation}

\begin{figure*}[t]
    \centering
    \includegraphics[width=0.8\textwidth]{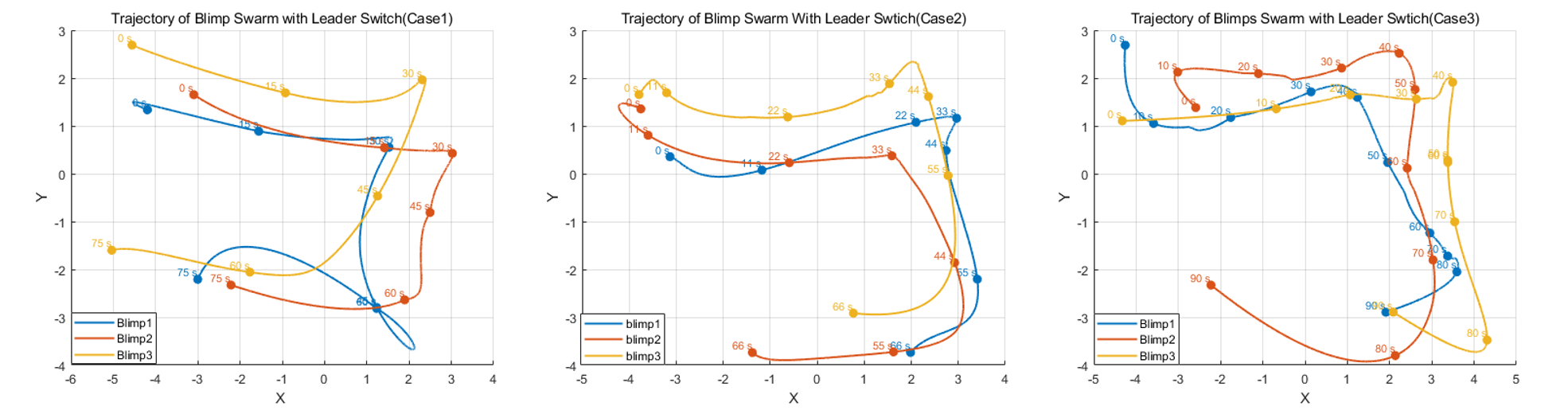} 
    \caption{Top-view trajectory of a blimp swarm executing two consecutive sharp turns with leader switch algorithm}
    \label{fig:success}
\end{figure*}
\begin{figure*}[t]
    \centering
    \includegraphics[width=0.8\textwidth]{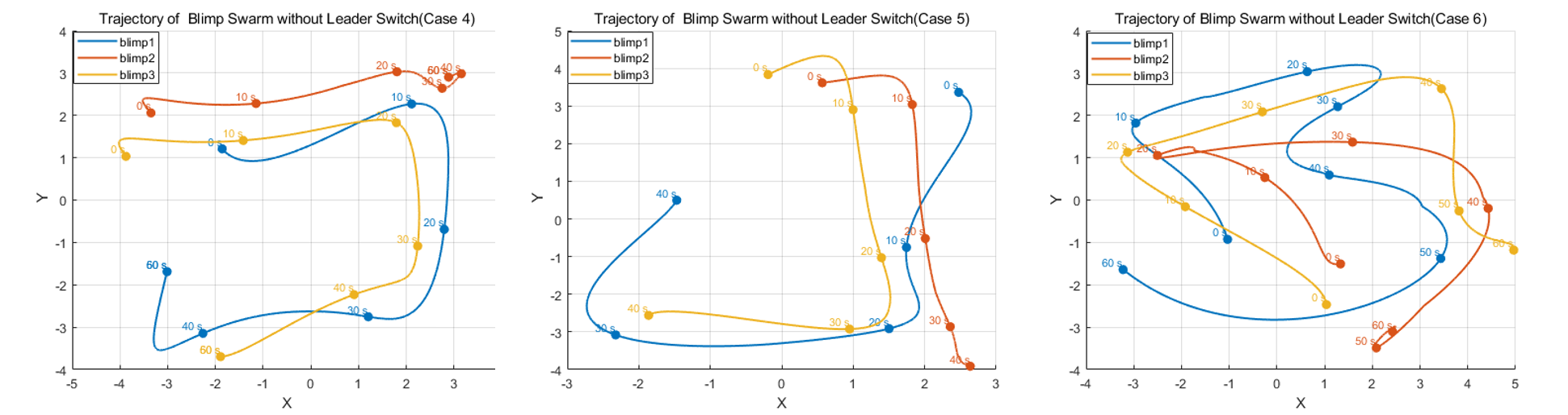} 
    \caption{Top-view trajectory of a blimp swarm executing two consecutive sharp turns without leader switch algorithm}
    \label{fig:fail}
\end{figure*}

\section{EXPERIMENT}
\subsection{Experimental Setup}
The experiment is designed to test a blimp swarm consisting of three blimps, starting from a designated starting point and following a predefined path with two sharp turns at waypoints j and j+1, leading to the goal point, as shown in Fig. \ref{fig:pic3}. The blimps are arranged in a leader-follower formation, where the leader blimp is manually controlled, and the followers adjust their positions based on visual feedback. The primary objective of the experiment is to evaluate the performance of the leader-switching mechanism during these sharp turns. The leader switch is triggered under specific maneuvering conditions, such as sharp acceleration or deviation in formation.

\subsection{Evaluation Metrics}
To demonstrate the generality of our algorithm, we set the initial formation to be random. This ensures the algorithm can handle diverse scenarios rather than being tailored to specific conditions. Given the variability in initial formations, the average area is used only as a reference, not a strict evaluation criterion. The experiment is primarily evaluated using:

\textbf{1. Root Mean Squared Error (RMSE)}: The RMSE of the triangle area formed by consecutive blimps is calculated before and after the sharp turns to assess formation stability. Lower RMSE values indicate better formation stability and minimal deviation from the desired shape during maneuvering.

\textbf{2. Success Rate of Turn Completion}: This metric monitors whether the blimp formation successfully completes the sharp turn without breaking. A successful turn is defined as all blimps maintaining their relative positions to the leader throughout the maneuver, with no separation from the formation.




\subsection{Results}
As illustrated in Fig. \ref{fig:success} and Fig.\ref{fig:fail}, we conducted six test cases to assess the impact of the leader-switching mechanism on the swarm's trajectory. The experiments were divided into two groups: Cases 1-3 utilized the leader-switching algorithm, while Cases 4-6 did not.

The results clearly indicate that the leader-switching mechanism significantly improves the swarm's ability to maintain formation during sharp turns. In the cases where the leader switch is employed (Cases 1-3), the experiments achieve a 100\% success rate, with all blimps completing the maneuver successfully. Additionally, these cases exhibit relatively low area RMSE values (ranging from 0.14 to 0.28), indicating high stability.

In contrast, the experiments without the leader switch (Cases 4-6) show a notable decline in performance. Only one experiment (Case 6) is successfully completed, resulting in a 33.3\% success rate. The cases without the leader switch also display significantly higher area RMSE values (ranging from 2.06 to 2.56), suggesting reduced formation stability and accuracy.



\begin{table}[t]
\centering
\small
\caption{Experiment Results}
\label{table:experiment_results}
\begin{tabular}{lccc}
\toprule
\textbf{Test Case} & \textbf{Completed} & \textbf{Average Area} & \textbf{Area RMSE} \\
\midrule
Case 1  &      Yes      &      1.28       &      0.14     \\
Case 2  &      Yes      &      1.04       &      0.28     \\
Case 3  &      Yes      &      1.01       &      0.21     \\
Case 4  &      No       &      3.33       &      2.44      \\
Case 5  &      No       &      2.28       &      2.56      \\
Case 6  &      Yes      &      2.37       &      2.06      \\
\bottomrule
\end{tabular}
\end{table}

\subsection{Discussion}
The results in Table \ref{table:experiment_results} demonstrate the effectiveness of the leader-switching mechanism in maintaining formation stability during sharp maneuvers. Compared to cases without leader switching, dynamic leader-switching significantly improved success rates and reduced RMSE values, highlighting its role in enhancing trajectory control precision and swarm coordination. This approach not only strengthens formation integrity but also reduces the operator's workload by minimizing manual corrections during complex maneuvers.

However, this method has limitations as the number of followers increases. With around 10 followers, visual occlusions among followers can occur, disrupting tracking and formation stability. A potential solution is a hierarchical following strategy, where some followers track the leader and others follow these followers in a 1-to-1 manner, helping to reduce occlusions and maintain stability in larger formations.

\section{CONCLUSIONS AND FUTURE WORK}
In this paper, we proposed an algorithm that leverages a leader switch to enhance the trajectory control of a blimp swarm. The experimental results demonstrate that our algorithm, when incorporating the leader switch, significantly improves the accuracy and stability of the swarm's movement. Specifically, the cases with the leader switch showed lower average area and area RMSE values compared to the cases without it, highlighting the effectiveness of our algorithm in ensuring coordinated and precise navigation in multi-agent systems.

As for future work, we plan to extend our algorithm to address challenges related to occlusions between followers in the swarm. 






\bibliographystyle{IEEEtran}
\bibliography{my} 

\end{document}